\RequirePackage{fix-cm}
\documentclass{svjour3}                     % onecolumn (standard format)
\smartqed  % flush right qed marks, e.g. at end of proof

\usepackage[utf8]{inputenc}
\usepackage{titling}
\usepackage[T1]{fontenc}
\usepackage{color}
\usepackage{appendix}
\usepackage{subfigure}
\usepackage{amssymb}
\usepackage{float}
\usepackage{xspace}
\usepackage{graphicx}
\usepackage{amsmath,xparse}
\usepackage{mathtools}
\usepackage{tabularx}
\usepackage{booktabs}
\usepackage{mathtools}
\usepackage{soul}
\usepackage[colorlinks = true,
            linkcolor = blue,
            urlcolor  = magenta,
            citecolor = blue,
            anchorcolor = blue]{hyperref}

\newlength{\cellwidth}
\usepackage{authblk}
\usepackage{latexsym}
\begin{document}

\title{A fast and memory-efficient algorithm for smooth interpolation of polyrigid transformations: application to human joint tracking}

\author[1,2]{Karim Makki}
\author[4]{Bhushan Borotikar}
\author[5]{Marc Garetier}
\author[4]{Sylvain Brochard}
\author[3]{Douraied Ben Salem}
\author[1]{François Rousseau}

\affil[1]{Institut Mines Télécom Atlantique, LaTIM U1101 INSERM, UBL, Brest, France.}
\affil[2]{Aix Marseille Univ, LIS UMR 7020 CNRS, Marseille, France.}
\affil[3]{Department of Neuroradiology, Forensic Imaging Unit, H. La Cavale Blanche, CHRU of Brest, LaTIM U1101 INSERM, Brest, France.}
\affil[4]{University of Western Brittany, LaTIM U1101 INSERM, Brest, France.}
\affil[5]{Department of Radiology, Military Teaching Hospital Clermont-Tonnerre, LaTIM U1101 INSERM, Brest, France.}

\date{}
\maketitle

\begin{abstract}
The log Euclidean polyrigid registration framework provides a way to smoothly estimate and interpolate poly-rigid/affine transformations for which the invertibility is guaranteed. This powerful and flexible mathematical framework is currently being used to track the human joint dynamics by first imposing bone rigidity constraints in order to synthetize the spatio-temporal joint deformations later. However, since no closed-form exists, then a computationally expensive integration of ordinary differential equations (ODEs) is required to perform image registration using this framework. To tackle this problem, the exponential map for solving these ODEs is computed using the scaling and squaring method in the literature. In this paper, we propose an algorithm using a matrix diagonalization based method for smooth interpolation of homogeneous polyrigid transformations of human joints during motion. The use of this alternative computational approach to integrate ODEs is well motivated by the fact that bone rigid transformations satisfy the mechanical constraints of human joint motion, which provide conditions that guarantee the diagonalizability of local bone transformations and consequently of the resulting joint transformations. In a comparison with the scaling and squaring method, we discuss the usefulness of the matrix eigendecomposition technique which reduces significantly the computational burden associated with the computation of matrix exponential over a dense regular grid. Finally, we have applied the method to enhance the temporal resolution of dynamic MRI sequences of the ankle joint. To conclude, numerical experiments show that the eigendecomposition method is more capable of balancing the trade-off between accuracy, computation time, and memory requirements. 
\end{abstract}

\keywords{Diffeomorphic registration \and Differential geometry \and Exponential map \and Lie theory \and Matrix eigendecomposition \and Motion interpolation \and ODEs}

\section{Introduction}
\label{sec:introduction}

Non-rigid or deformable registration is an important tool for assessing spatial and temporal changes between images~\cite{crum2004non,yu2019learning}. It allows a non-uniform mapping between the source and target images. Among these non-rigid registration techniques are the diffeomorphic ones which have presented some interesting properties such as invertibility, differentiability, and smoothness of the resulting deformation fields. Furthermore, these techniques are suitable for parameterizing the non-linear deformations of anatomical structures or shapes which include the aspect of change over time~\cite{arsigny2006log,durrleman2014morphometry}.
The Log Euclidean Polyrigid Framework (LEPF) is a parametric registration approach that composes a set of rigid transformations into a global diffeomorphism~\cite{arsigny2006log,arsigny2009fast}. The main idea is to parameterize a non-linear geometrical deformation with a small number of flexible degrees of freedom. Hence, it can be used to estimate human joint deformations by fusing a set of local bone rigid transformations, composing the joint of interest~\cite{makki2019vivo}. 
This approach, which is based on solving a first order \textit{autonomous} ODE modeling the temporal evolution of a dynamical system, offers the possibility to estimate \textit{infinitesimal diffeomorphisms}, parameterized by a dynamical time scale, via the exponential map from \textit{Lie Algebra} to the corresponding {Lie group}.
This framework maps the interpolation of a deformation field into the tangent space, and then transforms the resulting velocity-vector field back onto the manifold. Which makes it possible to estimate smooth deformation fields in continuous time where the \textit{invertibility} is guaranteed. The LEPF can be used to estimate a smooth and continuous curves of moving articulated structures by interpolating the estimated polyrigid joint transformations in the domain of matrix logarithms. Which is in fact a generalization of the parameterization of rigid body transformations via the exponential map, as presented in~\cite{vzefran1998interpolation,belta2002computation,beltanew,vemulapalli2014human}.\\
Since the LEPF relies on the computationally heavy solution of an ODE, the efficient computation of the exponential map over a regular grid requires the implementation of a fast algorithm to deal with the high-number of point trajectories to be estimated simultaneously. In this context, \textit{Arsigny et al.} have proposed to use a fast algorithm employing the scaling and squaring method~\cite{arsigny2009fast}. The major advantage of this technique is that it allows an accurate approximation of the exponential map regardless of whether the matrices  are diagonalizable or not. However, this method have a high memory requirement to store all matrices in the main computer memory during the repeated squarings.\\

Towards the end of the 1970s, \textit{Moler and Van Loan} have synthetized a study to present the different ways of computing the exponential of a square matrix~\cite{moler1978nineteen}. This study has been revisited more than two decades later in~\cite{moler2003nineteen} to take advantage of the advances of computational resources and to classify the existing methods and algorithms in terms of generality, accuracy, storage requirements, and efficiency. This study has pointed out that the algorithms which avoid use of the eigenvalues are more time-consuming for any particular problem.

Inspired by these conclusions, we propose to generate trajectories and motion interpolants from bone rigid transformations by scaling the transformation eigenvalues using a matrix-diagonalization based method. 

The matrix diagonalization method has been already used to interpolate $3\times 3$ symmetric positive-definite matrices (\textit{i.e.} tensor matrices) in~\cite{arsigny2007geometric}. A symmetric positive-definite $A$ matrix has always a spectral decomposition $A = QDQ^T$, where $Q$ is orthogonal and $D$ is diagonal with positive diagonal elements, and then $exp(t.log(A))$\\ $= Q.exp(t.log(D)).Q^T \quad _{t\in\{0,1\}}$ are also symmetric positive definite as they still representing infinetismal tensors.
% If A is nonsingular then it has at least 2^s square roots, where s is the number of distinct eigenvalues. The existence of a square root of a singular matrix depends on the Jordan structure of the zero eigenvalues.
However, this computational approach has never been used to interpolate $4 \times 4$ homogeneous transformations and the difficulty is to deal with these non symmetric matrices with nontrivial rotations generated by a linear dynamical system. In classical mechanics, one of the well known properties is that an homogeneous $4\times4$ transformation matrix cannot be diagonalized and can, at best, be reduced to a Jordan canonical form ~\cite{hoffman1971linear,mccarthy1990introduction}. Since these matrices are extensively used for robotics and computer vision applications, \textit{Ghosal et al.}~\cite{ghosal2006note} have tried to revolutionize these classical theories by demonstrating that these transformations can be diagonalized under a certain condition and the rule becomes the exception.   
% the diagonalizability of such transformations has never been discussed in the context of the LEPF before.   
In this work, we investigate the applicability of this technique for interpolating homogeneous bone rigid transformations and geometric transformations generated by a linear dynamical system.
The diagonalizability of bone transformations under joint mechanical contraints is justified in section~\ref{diagonalizability}. Then, we employ this technique to propose a fast algorithm for evaluating a Log-Euclidean polyrigid transformations on a regular grid. The main advantage of our algorithm is its capacity to drastically reduce memory requirements.
\\
To demonstrate the effectiveness of the proposed algorithm on both synthetic data and real dynamic MRI data, we performed an objective comparison between the two methods in terms of accuracy, computation time, and memory requirements. And we concluded that the eigendecomposition method is more capable of balancing the trade-off between them. 

Finally, we have applied the eigendecomposition technique to estimate a smooth continuous curve of the ankle joint trajectory from a discrete set of bone transformations. This allowed to interpolate intermediate time frames in-betweens acquired time frames and thus to enhance the temporal resolution of the dynamic MRI sequence.

To resume, the main contributions of this work are:
\begin{itemize}
    \item We define a trigonometric closed form for smooth parametrization of elements of $SE(3)$ without having to compute the matrix logarithms and exponentials in Section~\ref{closed_form}.
    \item We demonstrate the diagonalizability of human joint transformations in Section~\ref{diagonalizability} and we propose to compute the exponential map using the eigendecomposition technique with fewer matrix multiplications in Section~\ref{eigen}.
    \item We present an application of the proposed computational scheme in Section~\ref{dynmri}: spatio-temporal reconstruction of dynamic MRI sequences. 
\end{itemize}

\section{Methods}
\label{S:2}

\subsection{Smooth interpolation on $SE(3)$}

\subsubsection{Formulation of the interpolation problem}

Consider two elements $T_a$, $T_{b} \in SE(3)$. The interpolation function to estimate a smooth trajectory between them according to a time-varying parameter $t \in [0,1]$, is given by:
\begin{equation}
\gamma : SE(3) \times SE(3) \times \mathbb{R} \rightarrow SE(3)
\end{equation}
with:
\begin{equation}
    \gamma(T_a, T_b, 0) = T_a 
\end{equation}
\begin{equation}
    \gamma(T_a, T_b, 1) = T_b 
\end{equation}
\\ 
The group element $T_c \in SE(3)$ that takes $T_a$ to $T_b$ can be obtained as follows:

\begin{equation}\label{complex}
    T_c \equiv T_a^{-1} \circ T_b = T_b.T_a^{-1}
\end{equation}

The function $\gamma$ transforms the interpolation into the tangent space $se(3)$ for linearization purposes, and then transforms the resulting velocity vector back onto the manifold. A scaled map of $T_c$ to the Lie Algebra $se(3)$ (where $t$ is the scaling factor) is given by:
\begin{equation}
    T_c(t) = t.log(T_c)
\end{equation}
Finally, the exponential map allows for transforming back into the manifold, yielding the following interpolation function:
\begin{equation}
\label{interpolation_formulation}
    \gamma(T_a, T_b,t) = exp(T_c(t)).T_a
\end{equation}

\subsubsection{An ODE approach for the interpolation on $SE(3)$}
Suppose now that we have the initial condition $T_a = id$, then the displacement of the rigid body can be modeled by the following autonomous first order ODE~\cite{beltanew}:
\begin{equation}
    \dot{\gamma}(t)=\gamma(t).log(T_b)
\end{equation}
To simplify notations, we denote $T_b=T$. The solution of this ODE gives a smooth continuous curve of the rigid body trajectory (\textit{i.e.} its optimal geodesic curve between two rigid body poses):
\begin{equation}\label{algebra2group}
    \gamma(t) = exp(t.log(T)) \quad for \quad  t\in [0,1]
\end{equation}
The Equation~\eqref{algebra2group} shows that the exponential map takes the linear tangential trajectory $t.log(T) \in se(3)$, into a one-parameter subgroup of $SE(3)$, satisfying $\gamma(t_1+t_2)=\gamma(t_2).\gamma(t_1) $, for all $t_1, t_2$. This gives a local parameterization for the geometric transformation $T$.

Based on Equations~\eqref{complex},~\eqref{interpolation_formulation}, and~\eqref{algebra2group}, it is clear that the first formulation exhibits greater complexity than the ODE approach as it imposes additional matrix multiplications and inversion. 

\subsubsection{Smooth interpolation of polyrigid transformations}
To move from simply rigid to polyrigid interpolation of geometric transformations, Arsigny et al. proposed a registration framework to deal with autonomous continuous-time dynamical systems~\cite{arsigny2006log}. 
The main idea is to parameterize a non-linear geometrical deformation with a small number of intuitive parameters by fusing a set of linear transformations into a global diffeomorphism, according to a predefined positive weighting functions. 

Let $\mathcal{T}= \{T_i\}_{i \in \{1 \ldots N\}}$ be a set of $N$ transformations in $SE(3)$, then the associated Log-Euclidean Fréchet mean, representing the Riemannian equivalent of the Euclidean arithmetic mean, is obtained by minimizing the following metric dispersion:
\begin{equation}
    f_m(T_1,\ldots , T_N) = \underset{T}{argmin} \sum_{i=1}^{N} w_i d^2(T,T_i)
\end{equation}
where d(., .) is the distance associated to the Riemannian metric.

The resulting Log-Euclidean Fréchet mean is given by:

\begin{equation}
    f_m(T_1,\ldots , T_N)= exp(\sum_{i=1}^{N} w_i log(T_i))
\end{equation}

Finally, the polyrigid interpolation problem can be formulated as follows:

\textit{For a given point $x$ in the source image $D_{t_0}$, the continuous-time trajectory of $x$ from $D_{t_0}$ to the target image $D_{t_1}$ can be computed according to:}
\begin{equation}
\label{eq:log_euclid}
\Phi(\tilde{x}, t) = \exp \left(t\sum_{i=1}^{N}\tilde{w}_{i}(x) \log(\textrm{T}_{i})\right).\tilde{x}
\end{equation}
where $\tilde{x}=\begin{pmatrix}x\\1\end{pmatrix}$; $\Phi(\tilde{x},0)=\tilde{x}$; $t \in[0,1]$ is the time parameter; 
$\textrm{T}_{i} \in SE(3)$, is the rigid transformation of the component $i$ from $D_{t_0}$ to $D_{t_1}$; $N$ is the total number of rigid components; $\Phi(.,t)$ is the infinitesimal diffeomorphism from source to target; $\tilde{w}_{i}$ is a normalized weight function (\textit{i.e.}, $\sum_{i=1}^{N}\tilde{w}_{i}(x) = 1, \forall x \in D_{t_0}$) which defines the local influence of the $i^{th}$ component displacement on the resulting transformation at $x$ (see section~\ref{lepf}).

\subsubsection{Numerical computation of the exponential map}
The matrix exponential of an $m \times m$ matrix $T$ can be defined by the following convergent power series:
\begin{equation}
\label{eq:mat_expm}
e^T=\sum_{n=0}^{\infty}\frac{T^n}{n!}=I_{m}+T+\frac{T^2}{2!}+\frac{T^3}{3!}+\ldots
\end{equation}
where $I_m$ is the $m \times m$ identity matrix.

The computation of the exponential map for solving ODEs requires the development of a fast and efficient algorithm to deal with the repeated evaluation of the matrix exponential over a regular grid. In this context, \textit{Arsigny et al.} have proposed a fast algorithm for the estimation of exponential map to parameterize polyrigid transformations using the scaling and squaring method~\cite{arsigny2009fast}. The scaling and squaring method~\cite{al2009new,li2011matrix} is a recursive technique that exploits the fact that the matrix exponential can be easily estimated for matrices close to zero using the Pad\'e approximants. This method is based on the relation $e^T = (e^{\frac{T}{2^s}})^{2^s}$. The scaling step consists of evaluating $e^{\frac{T}{2^s}}$ while the squaring step consists of squaring the approximant $s$ times to finally obtain an estimation of $e^T$. The overall algorithm is composed of the three following steps:
\begin{itemize}
    \item \textbf{Scaling step:} $T\gets \frac{T}{2^s}$ so that $||T||_ {\infty}\simeq 1$, and $||T/2^s||\leq 1/2$.
    \item \textbf{Padé approximant to $e^T$:} $ r_m(T)=[m/m]$, with:
     $r_m(x) = p_m(x)/q_m(x)$; \\$p_m(x) = \sum_{i=1}^{m} \frac{(2m-i)!m!}{(2m)!(m-i)!} \frac{x^i}{i!} $ ; and $q_m(x) = p_m(-x)$, where $m$ represents the degree of the Pad\'e approximant. The resulting error satisfies:\\
     $e^x-r_m(x) = (-1)^m \frac{(m!)^2}{(2m)!(2m+1)!}x^{2m+1}+O(x^{2m+2})$.
    \item \textbf{Repeated squaring:} $e^T = (r_m(T))^{2^s}$.
\end{itemize}

In~\cite{arsigny2009fast}, the ODE integration is performed using the scaling and squaring method, based on the following equations:
\begin{equation}
    \gamma (.,t_j) = \left(\exp (\frac{1}{2^s}\sum_{i=1}^{N}\tilde{w}_{i}(.) \log(\textrm{T}_{i}))\right)^{t_j^{-1}}\label{halfway_squaring}
\end{equation}
\begin{equation}
\label{multiple_squaring}
    \gamma(.,1) = \gamma(.,\frac{1}{2^s})
\end{equation}
where $\gamma(.,0) = id$; $s$ is called the scaling factor, and $t_j^{-1} \in\{1,2\ldots 2^s\}$.\\
The efficiency of this technique over a regular grid is somehow comparable to that of the Fast Fourier Transform (FFT) according to Arsigny et al~\cite{arsigny2009fast}. However, this method increases the computation time dramatically with an important increase of grid size because of the high memory requirements for storing all matrices in the main computer memory during the repeated squaring.\\
For example, if $s=4$ in~\eqref{multiple_squaring}, $4$ matrix multiplications are needed to compute the exponential map at each grid point, as follows:
\[
\gamma(.,\frac{2}{16}) = \gamma(.,\frac{1}{16}) \circ \gamma(.,\frac{1}{16}),\]
\[\gamma(.,\frac{4}{16}) = \gamma(.,\frac{2}{16}) \circ \gamma(.,\frac{2}{16}),\]
\[\gamma(.,\frac{8}{16}) = \gamma(.,\frac{4}{16}) \circ \gamma(.,\frac{4}{16}),\] until we obtain: 

\[\gamma(.,1) = \gamma(.,\frac{8}{16}) \circ \gamma(.,\frac{8}{16}).\]

In general, the total number of matrix multiplications required for obtaining the complete trajectory of the articulated system is given by $N_s = 2^s = exp(s.log(2))$ so that the number $N_s$ increases exponentially with the scaling factor. From another angle, the increase of the scaling factor has been recommended in~\cite{higham2005scaling} for enhancing the numerical accuracy of matrix exponentials and their inverses. A review of Higham's method is proposed later in~\cite{higham2009scaling} for assessing the same accuracy with fewer matrix multiplications by increasing the degree of the Pad\'e approximant.
In~\cite{arsigny2009fast}, \textit{Arsigny et al.} recommended the use of a scaling factor $s \geq 6$ to obtain accurate and stable results.

\subsection{Proposed approach}
\label{contribution}

In this section, we define a closed form for smooth interpolation of simply rigid transformations, and then we investigate the diagonalizability of $4 \times 4$ homogeneous transformations and we propose to use a matrix diagonalization based method for fast computation of the exponential map with reduced  memory  consumption in the context of articulated registration.

\subsubsection{Closed form expression for smooth interpolation on $SE(3)$ via the exponential map:}
\label{closed_form}

Let $T \in SE(3)$, then $T$ has the form:

\begin{align}
T &= \begin{bmatrix*}
  \mathmakebox[\cellwidth]{[R]} & \mathmakebox[\cellwidth]{d} \\
  \textbf{0} & 1
\end{bmatrix*} 
\end{align}
where $[R] \in SO(3)$ is a $3 \times 3$ rotation matrix; $\textbf{0}$ is the $1 \times 3$ zero vector; and $d = (d_x,d_y,d_z)^T$ is a $3 \times 1$ translation vector. 

Let $\theta_x$,$\theta_y$, and $\theta_z \in ]-\pi,\pi]$ be the rotation angles performed by the rigid body about the x-axis, y-axis, and z-axis, respectively. Then for $\alpha_x = atan2(sin(\theta_x),$\\$cos(\theta_x))$; $\alpha_y = atan2(sin(\theta_y),cos(\theta_y))$; $\alpha_z = atan2(sin(\theta_z),cos(\theta_z))$; and $t \in [0,1]$, we define the following closed form for smooth parameterization of $T$ via the exponential map, without having to compute the matrix logarithm and exponential: 

\[
  \mathbf{e^{(t.log(T))}} =
  \begin{pmatrix}
   \begin{smallmatrix}
    cos(t. \alpha_y).cos(t. \alpha_z)
    \end{smallmatrix} & \begin{smallmatrix}
    -cos(t. \alpha_y).sin(t. \alpha_z) \end{smallmatrix}&
    \begin{smallmatrix}
   sin(t. \alpha_y) \end{smallmatrix} &
   \begin{smallmatrix}
    t.d_x\end{smallmatrix}
    \\
    \\
    \begin{smallmatrix}
    sin(t. \alpha_x).sin(t. \alpha_y).cos(t. \alpha_z)+ {} \\
     cos(t. \alpha_x).sin(t. \alpha_z) \end{smallmatrix} &
    \begin{smallmatrix}
    -sin(t. \alpha_x).sin(t. \alpha_y).sin(t. \alpha_z)+ {} \\
     cos(t. \alpha_x).cos(t. \alpha_z) \end{smallmatrix} &
    \begin{smallmatrix}
    -sin(t. \alpha_x).cos(t. \alpha_y)
    \end{smallmatrix} &
    \begin{smallmatrix}
    t.d_y
    \end{smallmatrix}
    \\
    \\
    \begin{smallmatrix}
    -cos(t. \alpha_x).sin(t. \alpha_y).cos(t. \alpha_z)+ {} \\
     sin(t. \alpha_x).sin(t. \alpha_z) \end{smallmatrix} &
    \begin{smallmatrix}
    cos(t. \alpha_x).sin(t. \alpha_y).sin(t. \alpha_z)+ {} \\
     sin(t. \alpha_x).cos(t. \alpha_z) \end{smallmatrix} &
    \begin{smallmatrix}
     cos(t. \alpha_x).cos(t. \alpha_y) \end{smallmatrix} &
    \begin{smallmatrix}
    t.d_z
    \end{smallmatrix}
    \\
    \\
    \begin{smallmatrix}
    0
    \end{smallmatrix} &
    \begin{smallmatrix}
    0
    \end{smallmatrix} &
    \begin{smallmatrix}
    0
    \end{smallmatrix} &
   \begin{smallmatrix}
    1
    \end{smallmatrix}
  \end{pmatrix}
\]
proof:

Let $R_x$, $R_y$ and $R_z$ be the three rotation matrices performed by the rigid body, about each direction in the space, then we have:
\begin{align}
R_x &= \begin{bmatrix*}
  \mathmakebox[\cellwidth]{1} & \mathmakebox[\cellwidth]{0} & \mathmakebox[\cellwidth]{0} \\
  0 & cos(\theta_x) & -sin(\theta_x)\\
  0 & sin(\theta_x) & cos(\theta_x)  
\end{bmatrix*} \\
R_y &= \begin{bmatrix*}
  \mathmakebox[\cellwidth]{cos(\theta_y)} & \mathmakebox[\cellwidth]{0} & \mathmakebox[\cellwidth]{sin(\theta_y)} \\
  0 & 1 & 0\\
  -sin(\theta_y) & 0 & cos(\theta_y)  
\end{bmatrix*} \\
R_z &= \begin{bmatrix*}
  \mathmakebox[\cellwidth]{cos(\theta_z)} & \mathmakebox[\cellwidth]{-sin(\theta_z)} & \mathmakebox[\cellwidth]{0} \\
  sin(\theta_z) & cos(\theta_z) & 0\\
  0 & 0 & 1 
\end{bmatrix*} 
\end{align}

For $(\theta_x,\theta_y,\theta_z) \in ]-\pi,\pi]^3$, one can map each rotation matrix to its tangent space $so(3)$ using the trigonometric function $atan2$. This gives:

\begin{align}
log(R_x) &= \begin{bmatrix*}
  \mathmakebox[\cellwidth]{0} & \mathmakebox[\cellwidth]{0} & \mathmakebox[\cellwidth]{0} \\
  0 & 0 & -\alpha_x\\
  0 & \alpha_x & 0  
\end{bmatrix*} \\
log(R_y) &= \begin{bmatrix*}
  \mathmakebox[\cellwidth]{0} & \mathmakebox[\cellwidth]{0} & \mathmakebox[\cellwidth]{\alpha_y} \\
  0 & 0 & 0\\
  -\alpha_y & 0 & 0  
\end{bmatrix*} \\
log(R_z) &= \begin{bmatrix*}
  \mathmakebox[\cellwidth]{0} & \mathmakebox[\cellwidth]{-\alpha_z} & \mathmakebox[\cellwidth]{0} \\
  \alpha_z & 0 & 0\\
  0 & 0 & 0 
\end{bmatrix*} 
\end{align}

Then, we define the explicit forms for transforming back these transformations into the manifold $SO(3)$ as follows:

\begin{align}
exp(t.log(R_x)) &= \begin{bmatrix*}
  \mathmakebox[\cellwidth]{1} & \mathmakebox[\cellwidth]{0} & \mathmakebox[\cellwidth]{0} \\
  0 & cos(t.\alpha_x) & -sin(t.\alpha_x)\\
  0 & sin(t.\alpha_x) & cos(t.\alpha_x)  
\end{bmatrix*} \\
exp(t.log(R_y)) &= \begin{bmatrix*}
  \mathmakebox[\cellwidth]{cos(t.\alpha_y)} & \mathmakebox[\cellwidth]{0} & \mathmakebox[\cellwidth]{sin(t.\alpha_y)} \\
  0 & 1 & 0\\
  -sin(t.\alpha_y) & 0 & cos(t.\alpha_y)  
\end{bmatrix*} \\
exp(t.log(R_z)) &= \begin{bmatrix*}
  \mathmakebox[\cellwidth]{cos(t.\alpha_z)} & \mathmakebox[\cellwidth]{-sin(t.\alpha_z)} & \mathmakebox[\cellwidth]{0} \\
  sin(t.\alpha_z) & cos(t.\alpha_z) & 0\\
  0 & 0 & 1 
\end{bmatrix*} 
\end{align}
Under the operation of composition, we obtain the exponential map associated to the rotation matrix $[R]$:
\begin{equation}
   exp(t.log(R)) = exp(t.log(R_x)).exp(t.log(R_y)).exp(t.log(R_z)) 
\end{equation}

And finally, we obtain the desired result by simply scaling the translation vector:

\begin{align}
exp(t.log(T)) &= \begin{bmatrix*}
  [exp(t.log(R))] & \mathmakebox[\cellwidth]{t.d} \\
  \textbf{0} & 1
\end{bmatrix*} 
\end{align}

However, and as mentionned from the beginning, one cannot extend these closed forms to the parameterization of polyrigid transformations via the exponential map.

Proof:

The log Euclidean mean of $N$ rigid transformations is given by: 

\begin{equation}
\label{eq:existance}
\mathcal{T}(.,t) = \exp \left(t\sum_{i=1}^{N}\tilde{w}_{i}(.) \log(\textrm{T}_{i})\right)
\end{equation}

We defined above a trigonometric closed form for smooth interpolation of each individual element of the special Euclidean group $SE(3)$. Since the elements of $SE(3)$ are not necessarily commutative (\textit{i.e.} $T_i.T_j \neq T_j.T_i$ for $i \neq j$ and $T_i, T_j \in SE(3)$), then the property $e^{(T_i+T_j)}=e^{T_i}.e^{T_j}$ is not verified and one cannot extend the proposed closed form to the polyrigid case.

\subsubsection{Diagonalizability of
the $4\times 4$ homogeneous transformation matrices}
\label{diagonalizability}

\begin{definition}
Let $T$ be a linear operator on the finite dimensional space $V$, then $T$ is diagonalizable if there is a basis for $V$ each vector of which is a charactarestic vector (eigenvector) of $T$. One could also define $T$ to be diagonalizable when the characteristic vectors of $T$ span $V$.
\end{definition}

% \begin{theorem}
% \hl{Let $T \in \mathbb{C}^n$ be a linear operator on the finite dimensional space $V$, let $\{\lambda_1..\lambda_n\}$ be teir , then $T$ is diagonalizable if and only if the algebraic multiplicity of each  be their eigenvalues, .}
% \end{theorem}

In classical mechanics, one of the well known properties is that an homogeneous $4\times4$ transformation matrix cannot be diagonalized and can, at best, be reduced to a Jordan canonical form ~\cite{mccarthy1990introduction}. If we take the example of screw motion through the $z-axis$, the screw transformation will be:

\begin{equation}
%\scriptstyle
T_{\theta}=  \begin{bmatrix}
cos(\theta) & -sin(\theta) & 0 & d_x \\
sin(\theta) & cos(\theta) & 0 & d_y \\
0 & 0 & 1 & d_z \\ 
0 & 0 & 0 & 1
\end{bmatrix} 
\end{equation}
where $T_{\theta} \in SE(3) $, $d=(d_x,d_y,d_z)$ is the translation vector, and $\theta$ is the rotation angle about the $z-axis$. In general, a rigid body displacement has a fixed line, called the screw axis. This axis is an invariant subspace of $T_{\theta}$ (\textit{i.e.} it has the same spatial position before and after applying the transformation to the rigid body).Consider the eigenvalue problem $T_{\theta}v= \lambda v$, the associated characteristic equation is $det(T_{\theta}-\lambda I_4)= (\lambda ^2 -2\lambda cos(\theta)+1) (1-\lambda) ^2=0$. Thus, the eigenvalues of $T_{\theta}$ in the complex domain are: $\{e^{i \theta}, e^{-i \theta}, 1, 1 \}$. The pair of complex conjugate $\lambda_1 = e^{i \theta}$ and $\lambda_2 = \lambda_1^* = e^{-i \theta}$ are the eigenvalues of the rotational block and the corresponding  eigenvectors $v_1=(x_1,0)$ and $v_2=(x_2,0) \in \mathbb{C}^4$ are four dimensional. Clearly, the eigenvectors associated with the repeated eigenvalue $\lambda_3 = \lambda_4 =1$ satisfy $[T-I_4]v = 0$. By definition, the eigenvector $v_3$ associated to $\lambda_3$ is the one derived from the Euler-Rodrigues' vector $b$, $v_3 = (b,0) =(tan(\frac{\theta}{2}).s,0)$ for $\theta \neq \{-\pi,\pi\}$, where $s$ is the unit vector in along the screw axis. So that the eigenvectors corresponding to the double root $\lambda =1$ are linearly dependant (\textit{i.e.} the algebraic multiplicity of $\lambda =1$ is two, different from its geometric multiplicity =1). Though $T_{\theta}$ does have a two dimensional invariant plane associated with this repeated eigenvalue. In that case, $T_{\theta}$ cannot be diagonalized  and can ,at best, be reduced to a Jordan canonical form.% For more details, the reader is referred to the work of \textit{McCarthy}}~\cite{mccarthy1990introduction}.

In~\cite{ghosal2006note}, \textit{Ghosal et al.} proved that one can exclude non-pure-screw transformations from these assumptions. A screw transformation occurs when the motion consists of a rotation about a single axis with a non-zero displacement along the same axis. More specifically, a $4\times 4$ homogeneous matrix with nontrivial rotation can be diagonalized if and only if the translation along the screw axis is zero (\textit{e.g.} $T_{\theta}$ can be diagonalized if and only if the translation along the $z-axis$, $d_z=0$). Under these constraints, \textit{Ghosal et al.} demonstrated the existence and the non-uniqueness of a fourth eigenvector $\tilde{v}_4$ of the form $\tilde{v} _4 = (u,0)+\epsilon(0,0,0,1)$, linearly independent of ${v}_3= (tan(\frac{\theta}{2}).s,0)$, and satisfying $[R-I_3]u = -\epsilon d$ where $R$ is the $3 \times 3$ rotation submatrix of $T$ and $\epsilon \neq 0$. As a consequence, the four distinct vectors $(x_1,0)$,$(x_2,0)$,$(b,0)$, and $(u,\epsilon)$ form an eigenbasis of $\mathbb{C}^4$ and the matrix $T_{\theta}$ is diagonalizable. If the displacement along the screw axis is non zero, \textit{Ghosal et al.} presented a rigorous algebraic treatement proposing explicit expressions for the Jordan basis by means of a small perturbation of the original matrix near the screw axis. \\

In the context of articulated motion, the human bones may never be able to perform screw motions in normal conditions due to the natural joint mechanical constraints. This suggests that rigid bone transformations are always diagonalizable. Moreover, the entire joint complex (bones plus surrounding tissues) moves in harmony. This suggests that even the resulting joint polyrigid transformations must be diagonalizable as they cannot represent screw transformations. A numerical example is given in the Appendix~\ref{appendix1}.

% \begin{figure}
% \begin{center}
% \begin{tikzpicture}[scale=0.47]
% \begin{axis}[
%  view={-20}{-20},
%  axis line style = ultra thick,
%  axis lines=middle,
%  zmax=80,
%   xmax=2,
%   ymax=2,
%  height=12cm,
%  xtick=\empty,
%  ytick=\empty,
%  ztick=\empty,
%  clip=false,
%  x label style={at={(axis cs:2,0.051)},anchor=north},
%   xlabel={$y$},
%  y label style={at={(axis cs:0.05,2)},anchor=north},
%   ylabel={$x$},
%  z label style={at={(axis cs:0.075,0,80)},anchor=north},
%   zlabel={$z$},
% ]
% \addplot3+[domain=0:11*pi,samples=500,samples y=0,orange,no marks,ultra thick] 
% ({sin(deg(x))}, 
% {cos(deg(x))}, 
% {6*x/(pi)})
% %node[circle,scale=0.2,fill,pos=0.05]{}
% %node [circle,scale=0.5,fill,pos=0.15]{} % coordinate (A)
% node [circle,scale=0.5,fill,pos=0.515]{}; % coordinate (B)
% %\draw (A)--(B);
% \end{axis}

% \end{tikzpicture}
% \caption{\label{fig:screw_motion} Example of screw motion through the z-axis.}
% \end{center}
% \end{figure}

\subsubsection{Smooth interpolation of polyrigid transformations via matrix eigendecomposition}
\label{eigen}
In the context of articulated registration, we propose to employ an eigenvalue-based method for smooth interpolation on $SE(3)$, assuming that the map from the Lie Algebra to the corresponding Lie group is performed according to the equation~\eqref{algebra2group} for each of the bones.

Let $\textrm{T} \in SE(3)$ be the rigid transform to be interpolated, then there exist an \textit{invertible} matrix $P$ and a diagonal matrix $D$ such that $T=PDP^{-1}$. The diagonal elements $\{\lambda_k\}_{k\in \{1\ldots4\}}$ are the eigenvalues of $T$  which satisfy the \textit{characteristic} equation $det(T-\lambda_k I_4)=0$, while the columns of $P$ are the corresponding eigenvectors $\{\textit{v}_k\}_{k\in \{1\ldots4\}}$ which satisfy the linear equation $T\textit{v}_k = \lambda_k \textit{v}_k$, also known as the \textit{eigenvalue problem}. Based on this matrix eigendecomposition, one can compute the rigid body trajectory with respect to~\eqref{algebra2group} as follows:

\begin{equation}
\label{eq:log_euclidean_rigid}
%\scriptstyle
 \gamma(t) = P. diag(e^{t.log(\lambda_1)},e^{t.log(\lambda_2)}, e^{t.log(\lambda_3)}, 1). P^{-1}
\end{equation}

\begin{definition}
A logarithm of $T \in \mathbb{C}^{n \times n}$ is any matrix $X$ such that $e^X=T$.
\end{definition}

The matrix logarithm of most practical interest is the one whose eigenvalues lie in the right half-plane, which is called the principal matrix logarithm. If $T$ is nonsingular (\textit{i.e.} if $T$ is invertible) and has no eigenvalues on the negative real axis then $T$ has a unique principal matrix logarithm. For a diagonalizable matrix $T$, $log(T) = P. diag(log(\lambda_1),log(\lambda_2), log(\lambda_3), 0). P^{-1}$.

\begin{figure*}[t!]
\centering
\subfigure{\includegraphics[scale=0.15]{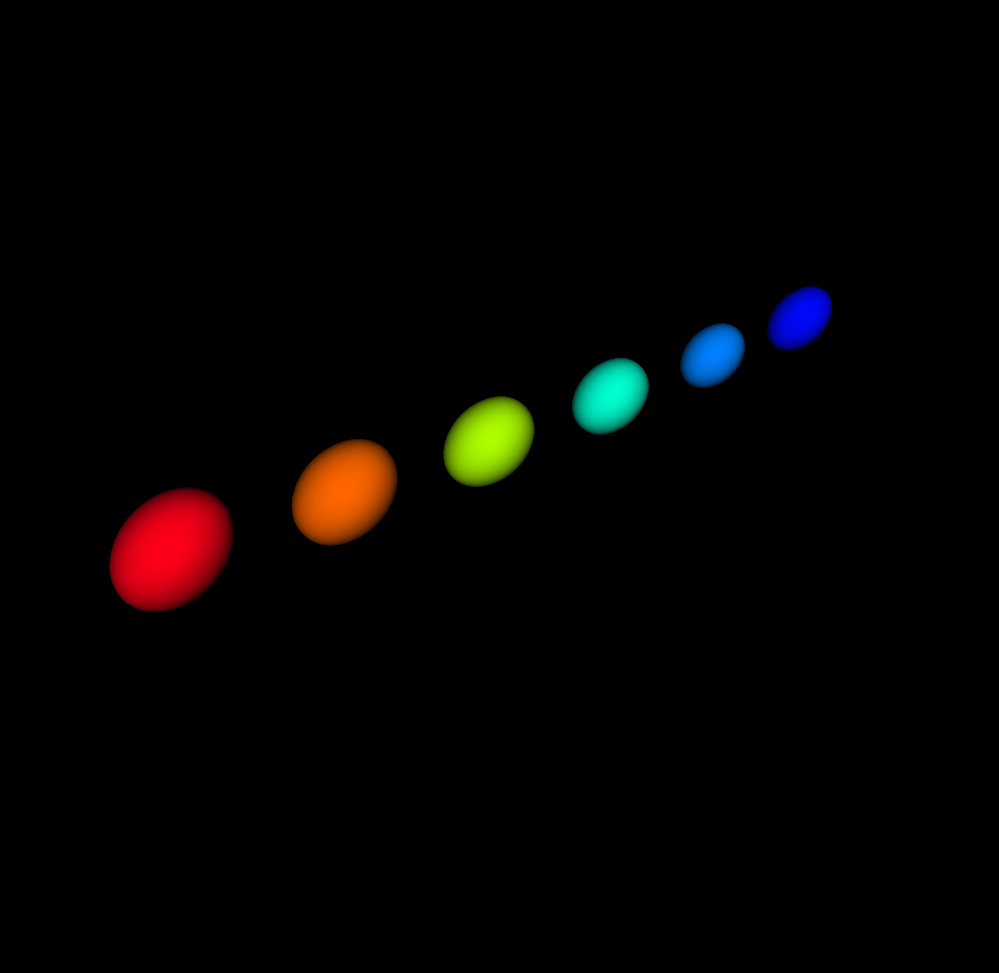}}
\subfigure{\includegraphics[scale=0.15]{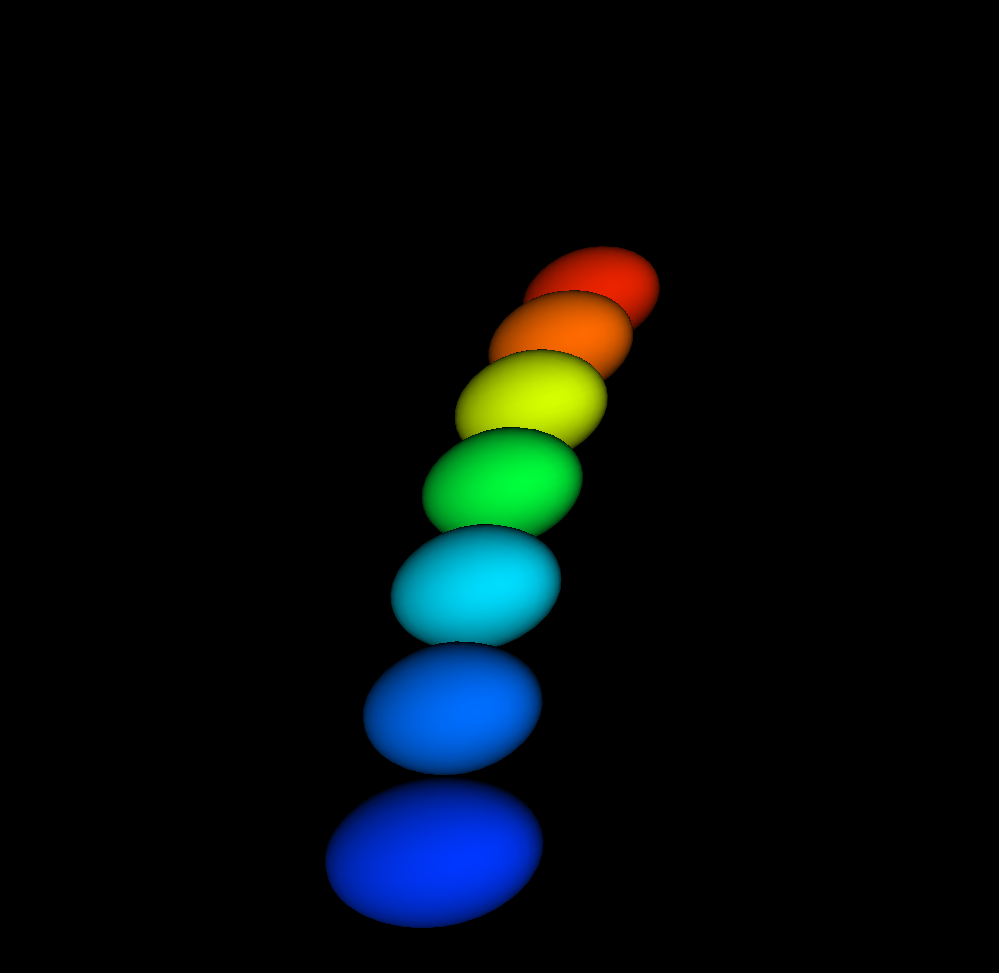}}
\hspace{0.05mm}
\subfigure{\includegraphics[height = 5.25cm,width=0.8cm]{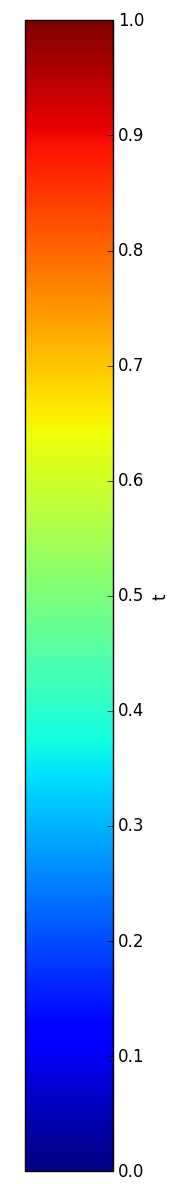}}
\caption{\label{fig:smooth_interp} Smooth interpolation of linear geometric transformations that map the blue ellipsoid (initial pose) to the red one (final pose), obtained by varying $t$ from $0$ to $1$ in~\eqref {eq:power}. From left to right: interpolation of a simulated affine transformation with the following parameters: translations: $(0,0,0mm)$, rotations: $(0,0,10^{\circ})$, shearings: $(0,0,0)$, and scalings: $(1.5,1.5,1.5)$; interpolation of a simulated rigid transformation (a single rotation of $30^{\circ}$ around the $z-axis$).}

\end{figure*}

Thanks to the property $exp(t.log(T))=T^{t}$, one can reformulate the interpolation function as follows:
\begin{equation}
\label{eq:power}
%\scriptstyle
\gamma({t})= T^{t} = PD^{t}P^{-1} = P. diag(\lambda_1^{t},\lambda_2^{t}, \lambda_3^{t}, 1). P^{-1}
\end{equation}
Changing $t$ continuously from $0$ to $1$ will \textit{infinitesimally} change the function $\gamma(t)$ from the \textit{identity} to the matrix $T$. This allows for interpolating between two rigid body poses from an homogeneous geometric transformation, by directly scaling its eigenvalues where $t$ is the scaling factor here. Figure~\ref{fig:smooth_interp} illustrates the linear interpolation of one simulated 3D rigid transformation using~\eqref{eq:power}.\\
This technique can be also applied to interpolate any linear combination of bone rigid transformations and can thereby be embedded in the LEPF in this context.

\begin{figure*}[t!]
\centering
%\begin{minipage}{0.36\textwidth}
\subfigure{\includegraphics[scale=0.359]{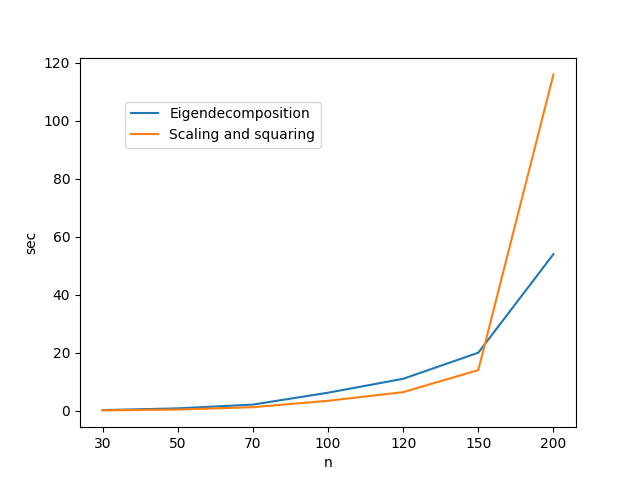}}
\subfigure{\includegraphics[scale=0.359]{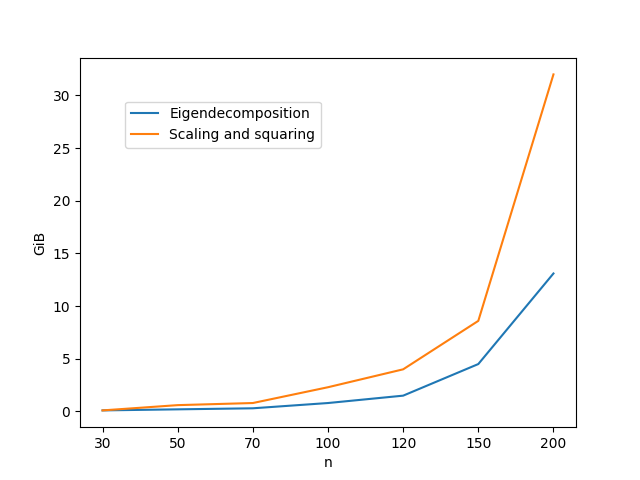}}
\caption{\label{computation_time} Computation of the exponential map over an $n \times n \times n$ regular grid. From left to right: computation times, and memory requirements in function of the grid size.}
\end{figure*}

\section{Applications and results}
\label{apps}

\subsection{Exponential map and parameterization of pure rotations: example of counterclockwise rotations}
\label{pure_rotations}
Let $T_{\theta}$ be the following $3\times 3$ matrix,    
%\begin{equation*}
\begin{center}
\[
%\resizebox{.2\textwidth}{!}{ 
T_{\theta} =
\begin{bmatrix}

 a & -b &  0 \\
 b & a &  0 \\
 0 & 0 &  c 
\end{bmatrix},
\]   %}
\end{center}

letting $a = cos(\theta)$; $b = sin(\theta)$; and $c = 1$, then the matrix $T_{\theta}$ is the homogeneous rigid transformation which generates a counterclockwise rotation by $\theta$ radians about the origin of a 2D cartesian coordinate system. The logarithm of the matrix $T_{\theta}$, which maps $T_{\theta}$ to its tangent space, is unique for $-\pi <\theta < \pi\quad rad$ and is equal to its \textit{principal matrix logarithm}:
\begin{center}
% \resizebox{.35\textwidth}{!}{ 
 \[
 log(T_{\theta})= \begin{bmatrix}
 0 & -atan2(b,a) & 0 \\
 atan2(b,a) & 0 & 0 \\
 0 & 0 & 0 
\end{bmatrix},
 \]
% }
\end{center}
a one-parameter subgroup of the Lie group $SE(2)$, which is parametrized with the parameter $t$, is given by: 

\begin{center}
%\resizebox{.48\textwidth}{!}{  
\[
 exp(t.log(T_{\theta}))  = \begin{bmatrix}  
 cos(t.atan2(b,a)) & -sin(t.atan2(b,a)) &  0 \\
 sin(t.atan2(b,a)) & \ cos(t.atan2(b,a)) &   0 \\
 0 & 0 & 1 
\end{bmatrix}.
\] % }
\end{center}

For $t=1$ and for different values of $\theta$, Table~\ref{comparison_table_exp_map} reports the error in the computation of the exponential map from the Lie algebra $se(2)$ to the corresponding Lie group $SE(2)$, using each of the two methods. This error is defined by: $\epsilon_{\theta} = \left \| exp(log(T_{\theta}))-T_{\theta} \right \|_2$.\\
Table~\ref{comparison_table_exp_map} shows that both methods are prone to neglectable rounding errors: occurring during the matrix inversion and multiplications for the eigendecomposition method; and during the repeated squarings for the scaling and squaring method. This confirms that both methods have nearly the same precision.

\begin{table*}
  \centering
  \caption{Errors on exponential map estimates $exp(log(T_{\theta}))$ ($\epsilon_{\theta}$ in the order of  $10^{-16}$). For the second method, the matrix logarithm is first approximated using an inverse scaling and squaring method~\cite{al2012improved} based on the fact that the principal matrix logarithm is much simpler to compute for matrices close to the \textit{identity}.}
  \begin{tabularx}{\textwidth}{@{}>{\bfseries}c*{9}{X}@{}}
  \toprule
  \textbf{Method}& \multicolumn{9}{c@{}}{Error per rotation angle $\theta$} \\
  \cmidrule(l){2-10} & $-\frac{\pi}{2}$ & $-\frac{\pi}{4}$ & $-\frac{\pi}{6}$ & $\frac{\pi}{6}$ & $\frac{\pi}{4}$ & $\frac{\pi}{3}$ & $\frac{\pi}{2}$ & $\frac{2\pi}{3}$ & $\frac{3\pi}{4}$ \\
  \cmidrule(r){1-1}\cmidrule(l){2-10}
  1. Eigendecomposition & $3$ & $2.4$ & $2.3$ & $2.3$ & $2.4$ & $2.4$ & $3.2$ & $2.5$ & $1.2$ \\
  
  2. Scaling and squaring & $3.9$ & $3.8$ & $2.2$ & $2.2$ & $3.8$ & $4.1$ & $4.6$ & $11$ & $1.9$ \\
%   3 & 0     & 0     & 0     & 0     & 0     & 0     & 0     & 0     & 0 \\
  \bottomrule
  \end{tabularx}%
  \label{comparison_table_exp_map}
\end{table*}%

In order to compare the performances of the two methods in terms of computation times, we have generated a set of regular grids with a size of $(n \times n \times n \times 4 \times 4)$, with $n \in \{30,50,70,100,120,150,200\}$. Each element of each grid is equal to  $V=log(T_{\frac{\pi}{4}})$, where $T_{\frac{\pi}{4}}$ is the homogeneous rigid transformation $\in SE(3)$ that encodes a counterclockwise rotation by $\frac{\pi}{4}$ radians around the $z-axis$. Our code is implemented in \textit{Python} using the LAPACK (Linear Algebra Package) routines for computing the exponential map: we have used the routine \textit{tensorflow.linalg.expm} which uses a combination of the scaling and squaring method and the Pad\'e approximation (this routine optimizes the scaling factor automatically in order to increase the computational accuracy, depending on the matrix norm; for example for matrices \textit{very close} to zero, the minimal possible scaling factor $2$ is used), and the routine \textit{numpy.linalg.eig} to compute the matrix eigenvalues and eigenvectors. All experiments are performed on an Intel$^{\mbox{\scriptsize{\textregistered}}}$  Xeon$^{\mbox{\scriptsize{\textregistered}}}$  Processor E3-1271 v3 3.60 GHz, with a physical memory of 32GB.

Both methods are fast and numerically stable for sparse regular grids (\textit{i.e.}, for $n \leq 100$). As illustrated in Figure~\ref{computation_time}, the eigendecomposition method is twice slower than the scaling and squaring method for sparse grids because of the
additional time required for matrix inversion and multiplications, after solving the \textit{eigenvalue problem}. However, the scaling and squaring method increased the computation time dramatically with an important increase of grid size (\textit{i.e.}, for $n=200$) because of the high memory requirements for storing all matrices in the main computer memory during the repeated squarings.

\subsection{Application to the enhancement of temporal resolution of dynamic MRI sequences}
\label{dynmri}
\subsubsection{Motivation}
Dynamic MRI is a \textit{non-invasive} imaging technique that allows to qualitatively assess the behavior of the human joints \textit{in vivo}. Real-time Fast Field Echo (FFE) sequences have the potential to reduce the effect of motion artifacts by acquiring the image data within a few milliseconds~\cite{schaeffter2001projection}. However, the short acquisition times affect the temporal resolution of the acquired sequences (\textit{i.e.} the scanning duration is short relative to the joint motion). In  this work, we propose to reconstruct the missing frames of the sequence given the reduced amount of acquired data, by applying the proposed motion interpolation technique which is based on scaling the transformation eigenvalues. This leads to estimate a continuous-time trajectory of the joint from a discrete set of rigid-body poses. To do this: we first estimate the rigid motion of each bone from the acquired discrete data using a robust intensity-based registration algorithm. Then, we fuse these local transformations into a global diffeomorphisms to compute the non-linear joint deformations between successive images according to~\eqref{eq:log_euclid}. The idea is to reconstruct the missing time frames by interpolating the joint velocity vector field and then transforming the resulting \textit{infinitesimal} velocity vector field back onto the manifold via the exponential map. The algorithm has been applied and validated using dynamic data from five subjects performing passive ankle dorsi-plantar flexion. 

\subsubsection{Estimation of temporal joint deformation field}
\label{lepf}
The joint deformation field between each successive acquired time frames is estimated by fusing a set of rigid body transformations $\{T^{t_j}_i\}_{i \in 1 \ldots N}^{j \in 1 \ldots L-1}$ (corresponding to the bones of interest, where $L$ is the length of the acquired discrete sequence), according to Eq{~\eqref{eq:log_euclid}}. A continuous trajectory of the joint is then obtained by varying the time parameter from $0$ to $1$. 

\begin{center}
\begin{figure*}[!htb]
\minipage{0.4\textwidth}
  \includegraphics[width=\linewidth]{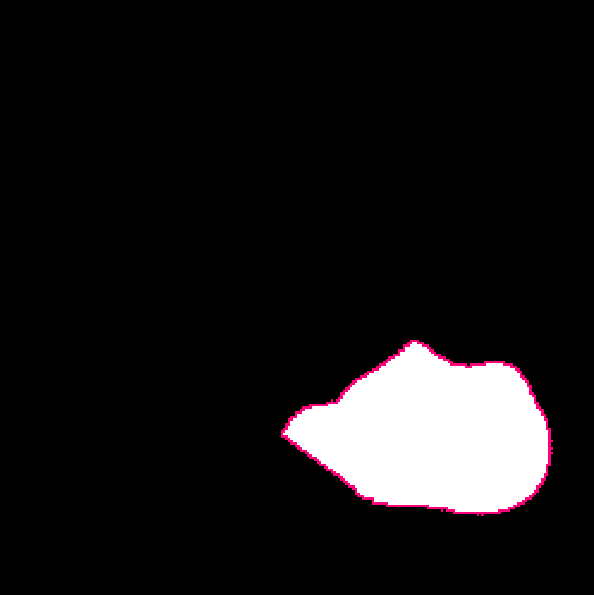}
  %\subcaption{}\label{fig:gt}
\endminipage\hfill
\minipage{0.4\textwidth}%
  \includegraphics[width=\linewidth]{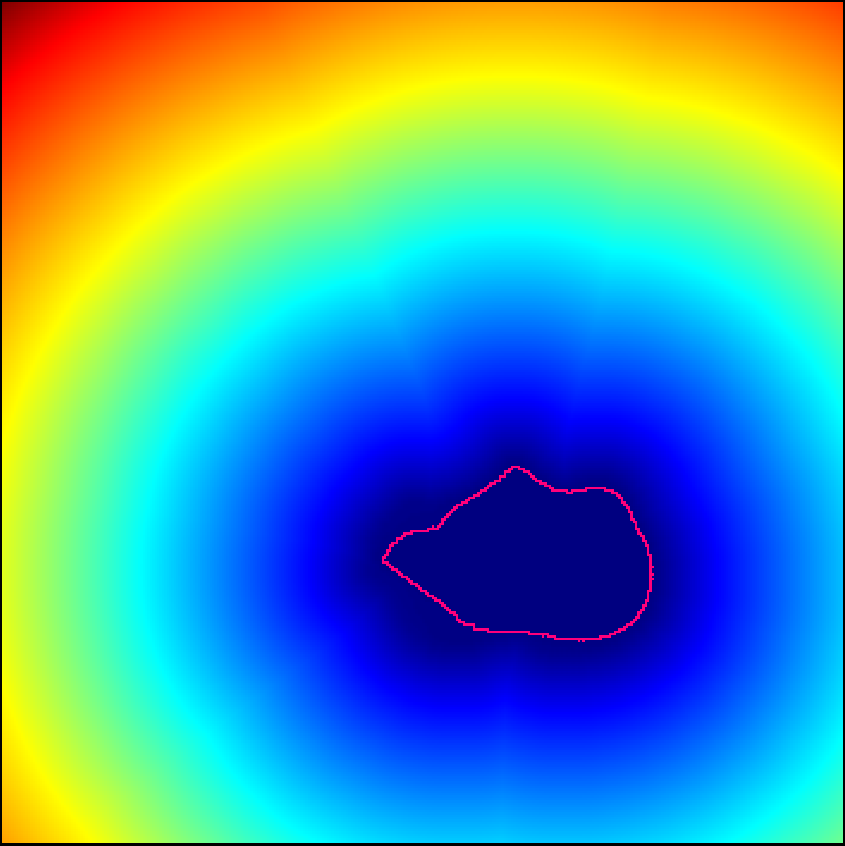}
  %\subcaption{}\label{fig:awesome_image3}
\endminipage\hfill
\minipage{0.18\textwidth}%
  \includegraphics[height= 5.13cm, width=1.3cm]{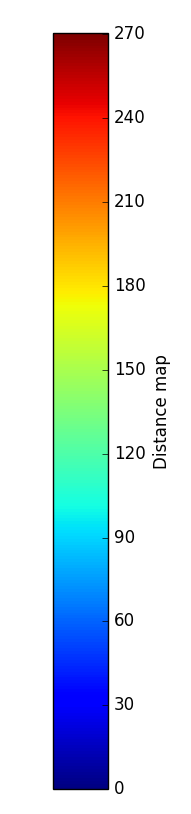}
  %\subcaption{Euclidean distance map of (b)}\label{fig:awesome_image3}
\endminipage
\caption{Euclidean distance map, from left to right: binary mask of the calcaneus; associated 3D Euclidean distance map.}
\label{fig:distance_map}
\end{figure*}
\end{center}

In~\cite{commowick2006efficient,commowick2008efficient}, \textit{Commowick et al.} proposed an inverse-distance based weighting function that preserves bone shapes after registration~\eqref{eq:weights111}. However, such weighting functions yield inaccurate deformation outside the segmented bones.
\begin{equation}
\label{eq:weights111}
w^{t_j}_{i}(x) = \frac{1}{1+\alpha.{dist(x,B^{{t_j}}_{i})^{\beta}}}
\end{equation}
where $x \in D^{t_j}$, $B^{{t_j}}_{i}$ is the binary mask of the $i^{th}$ component, and $dist(.,B^{{t_j}}_{i})$ is the corresponding Euclidean distance map at time $t_j$ (see Figure~\ref{fig:distance_map}). \\
As illustrated in Figure~\ref{fig:deformation_accuracy}, experiments show that an increase in $\beta$ in Eq~\eqref{eq:weights111} increases the accuracy of the estimated deformation field. Inspired by the relation ($e^x >> x^{\beta}$, when $x \rightarrow +\infty$), we have redefined the associated weight functions $\{w^{{t_j}}_{i}\}_{i \in 1\ldots N}$ as follows~\cite{makki2019temp}:
\begin{equation}
\label{eq:weights222}
w^{{t_j}}_{i}(x) = \frac{2}{1+\exp(\gamma. dist(x,B^{{t_j}}_{i}))}
\end{equation}
%  For more details about the definition of weighting functions, the reader is referred to~\cite{makki2019temp}.
To take into account the effects of all dependee components during fusion, each component weight function is normalized as follows: 
\begin{equation}
\tilde{w}^{{t_j}}_{i}=\frac{w^{{t_j}}_{i}}{\sum_{i=1}^{N} w^{{t_j}}_{i} }
\end{equation}

\begin{figure}[h!]
\centering
\begin{minipage}{0.42\textwidth}
\subfigure{\includegraphics[scale=0.2]{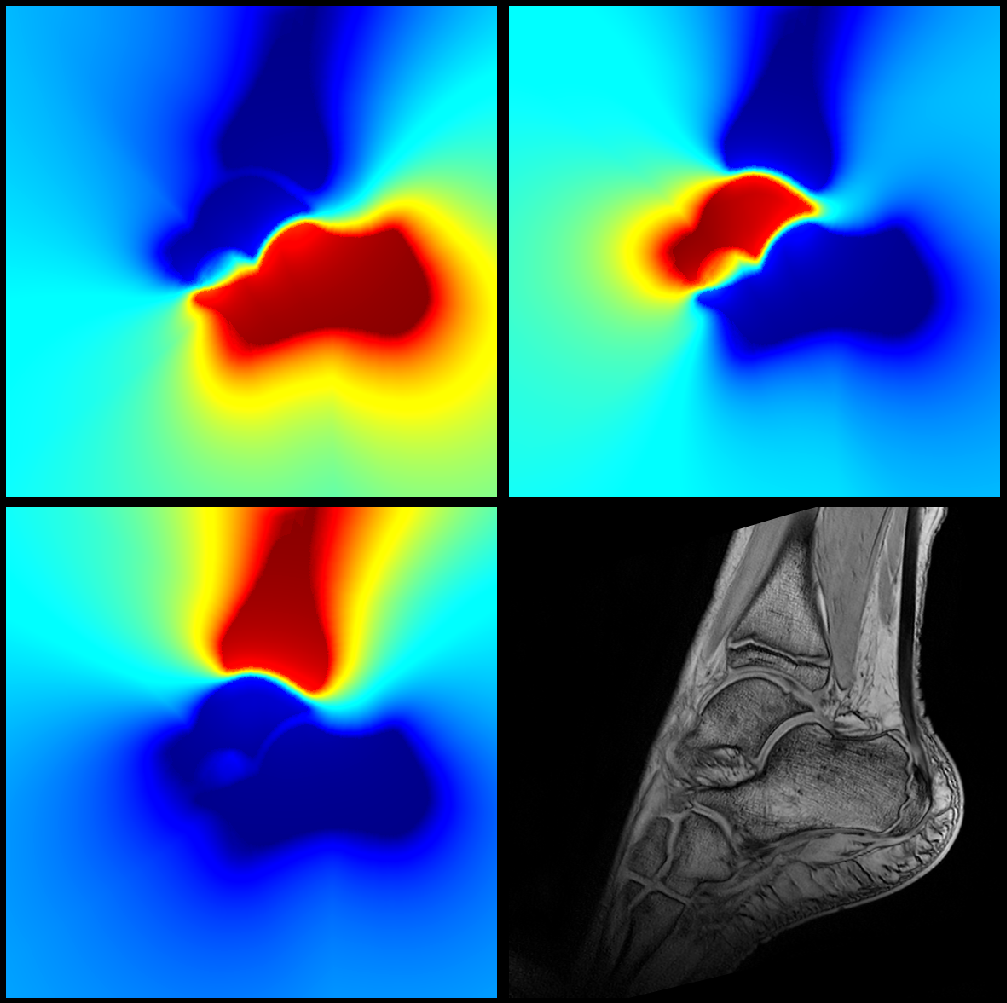}}
\end{minipage}
\hfill
\begin{minipage}{0.02\textwidth}
\subfigure{\includegraphics[height = 7.2cm,width=0.8cm]{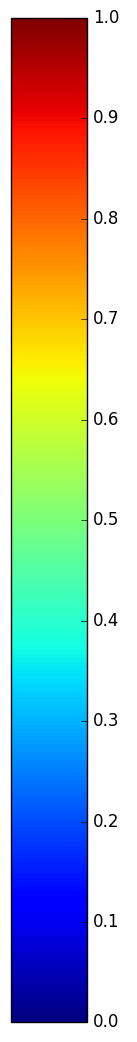}}
\end{minipage}
\caption{\label{fig:weight_functions} Normalized weighting functions: from up to down, from left to right: for the calcaneus; for the talus; for the tibia; and the associated MRI scan.}
\end{figure}

\begin{figure*}  [!t]

\begin{center}

  \includegraphics[width=12cm,height=8.3cm]{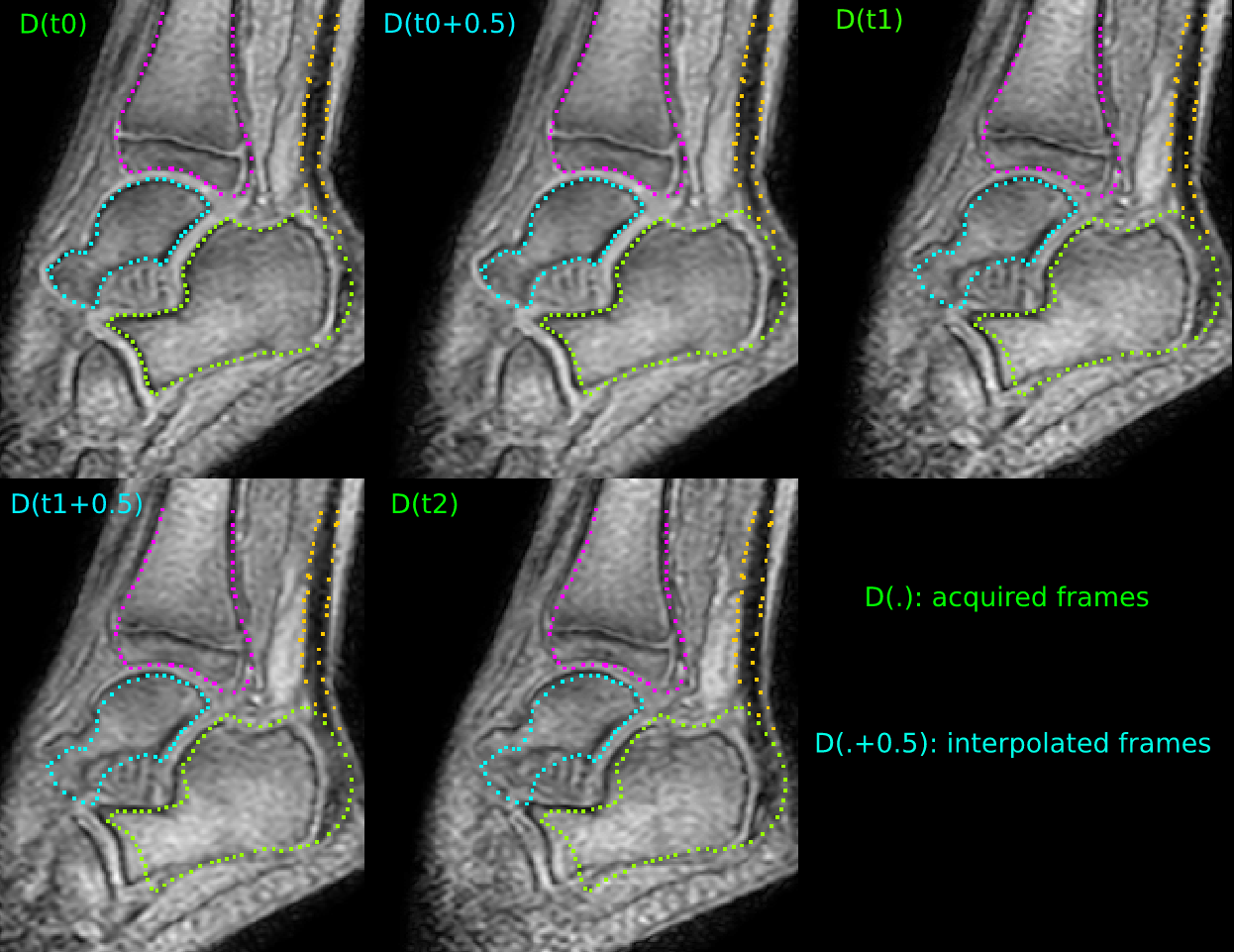}
\caption{Interpolation of missing time frames using the proposed eigendecomposition method: $D(t_j)$ is the $j^{th}$ acquired time frame; while $D(t_j+0.5)$ is the time frame half way between $D(t_j)$ and $D(t_{j+1})$ (\textit{i.e.}, $t =0.5$ in~\eqref{eq:log_euclid}), for $j \in \{0,1,2\}$. Bone contours have been drawn in the first time frame to show the interpolation quality and the relative joint motion.} 
\label{fig:enhancement}
\end{center} 
\end{figure*}

Figure~\ref{fig:weight_functions} illustrates each bone normalized weight function as defined in~\eqref{eq:weights222}.\\
Finally, the source image intensities are mapped to new coordinates in the target image space by spline interpolation. The robustness of the method and the accuracy of the results have been evaluated using a local leave-one-out cross-validation technique. All results on joint motion estimation and interpolation are validated and reported in~\cite{makki2019temp}.\\
The exponential map of Eq~\eqref{eq:log_euclid} is computed within $2.2$ seconds on a $352 \times 352 \times 6$ grid using the scaling and squaring method (which required 1.8 GB of RAM), and within nearly $4$ seconds using the proposed eigendecomposition method (which required 0.8 GB of RAM).  Figure~\ref{fig:enhancement} illustrates the interpolation quality of one 3D+time MRI sequence from our dataset.\\

\begin{figure*}[t!]
\centering
\subfigure{\includegraphics[scale=0.14]{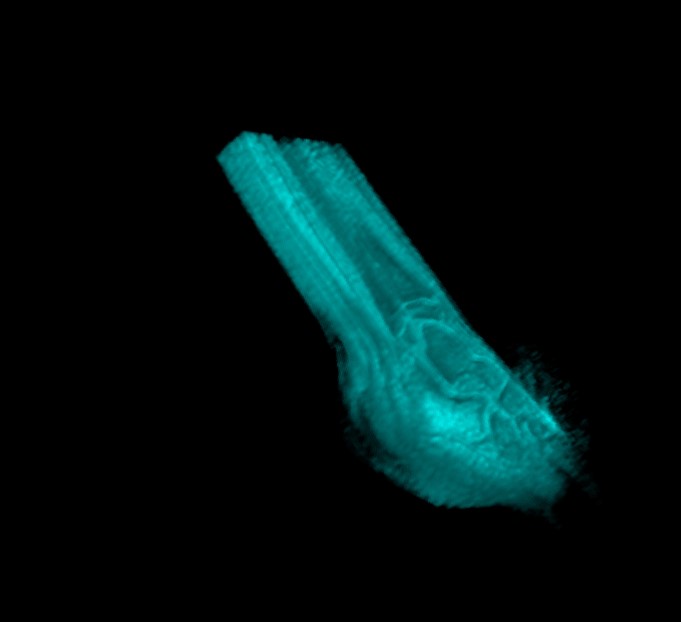}}
\subfigure{\includegraphics[scale=0.377]{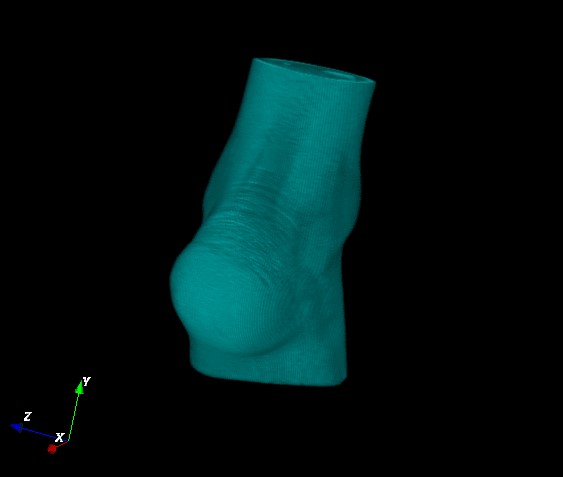}}
\caption{\label{reconstruction} High-resolution reconstruction of dynamic time frames. From left to right: the first low resolution acquired frame; and the corresponding volume, reconstructed in the same resolution of the static MRI scan.}
\end{figure*}

\subsection{Discussion}
If $m$ is the total number of intermediate points chosen to discretize the continuous trajectory of each point $x$, the scaling and squaring method is the most commonly used technique for computing the matrix exponential in previous works~\cite{arsigny2009fast,commowick2008efficient,ehrhardt2019temporal,porras2018locally}. However, the use of this method results in excessive memory requirements as it requires multiple matrix multiplications (see Figure ~\ref{computation_time}) and thus it is not suitable for estimating or interpolating very dense deformation fields (\textit{e.g.} when registering very high-resolution MRI data~\cite{makki2018high}). To overcome this limitation, we have proposed to compute the matrix exponential using matrix eigendecomposition. The grid is interpreted as a tensor of $dim_0\times dim_1\times dim_2$ matrices, each of size $4\times4$ (with $dim_0$, $dim_1$, and $dim_2$ are the image dimensions), and the trajectories of all the points of the regular grid are computed simultaneously. Furthermore, computations are performed in the complex domain: for all real transformation matrices expressed in homogeneous coordinates, the associated eigenvalues occur in complex-conjugate pairs where the imaginary part of the eigenvalue is the frequency of voxel rotation.\\ 

Both methods lead to nearly the same level of accuracy and they can also reach the machine precision since the error is always of order $10^{-16}$.\\
For sparse regular grids, the eigendecomposition method is a little slower than the scaling and squaring method, but it offers much faster computations in the case of dense regular grids. Hence, this method is more capable of balancing the trade-off between computation times and memory requirements. For example, the optimization of large storage requirements could be very useful when estimating a dense deformation field from a high-resolution static MRI scan to a low-resolution dynamic time frame~\cite{makki2018high,makki2019vivo}. In these previous works, a deformation vector field of $576 \times 576 \times 202 \approx 67$ millions deformation vectors is computed in 15 min using the eigendecompostion method while consuming the entire physical memory (32GB). However, the use of the scaling and squaring method for estimating the same deformation vector field results severe memory swapping which drastically slows down computations until totally stopping the computation process. Figure~\ref{reconstruction} illustrates the high resolution reconstruction of the ankle joint by estimating a dense deformation field from the static scan to a low-resolution time frame from the MRI sequence.\\
The eigendecomposition method allows for interpolating the desirable number of intermediate points and thus for obtaining a smooth and continuous trajectory of the dynamical system by scaling the transformation eigenvalues, without the need to perform a large number of matrix multiplications. In fact, once the \textit{eigenvalue problem} is solved over the regular grid, the transformation eigenvalues and the sub-matrices containing the associated eigenvectors (with their inverses) can be stored in the main computer memory, and the interpolation of an intermediate point requires only two matrix multiplications.

The proposed interpolation technique can be also extended to diagonalizable \textit{polyaffine} transformations, for which local transformations to be fused offers additional degrees of freedom like the scalings and shearings. Figure~\ref{fig:smooth_interp} illustrates the consistent interpolation of a simulated affine transformation using matrix eigendecomposition.

In most applications such as human skeleton tracking during motion, the amount of local rotation of each rigid component is strictly below $\pi$ radians, so that the matrix logarithm of a bone transformation $T$ exists always and is equal to the \textit{principal matrix logarithm}. This suggests that the property $exp(t.log(T))=T^{t}$, is verified for these transformations. Concerning the diagonalizability of $4 \times 4$ homogeneous rigid transformations, we argue that the joint bones may never be able to perform screw motions in normal conditions due to the effect of mechanical constraints on the joint motion. Hence, the estimated bone transformations expressed in homogeneous coordinates possess 4 linearly independent eigenvectors and thus are always \textit{diagonalizable}. \\

\begin{figure}[h!]
\centering

\includegraphics[scale=0.245]{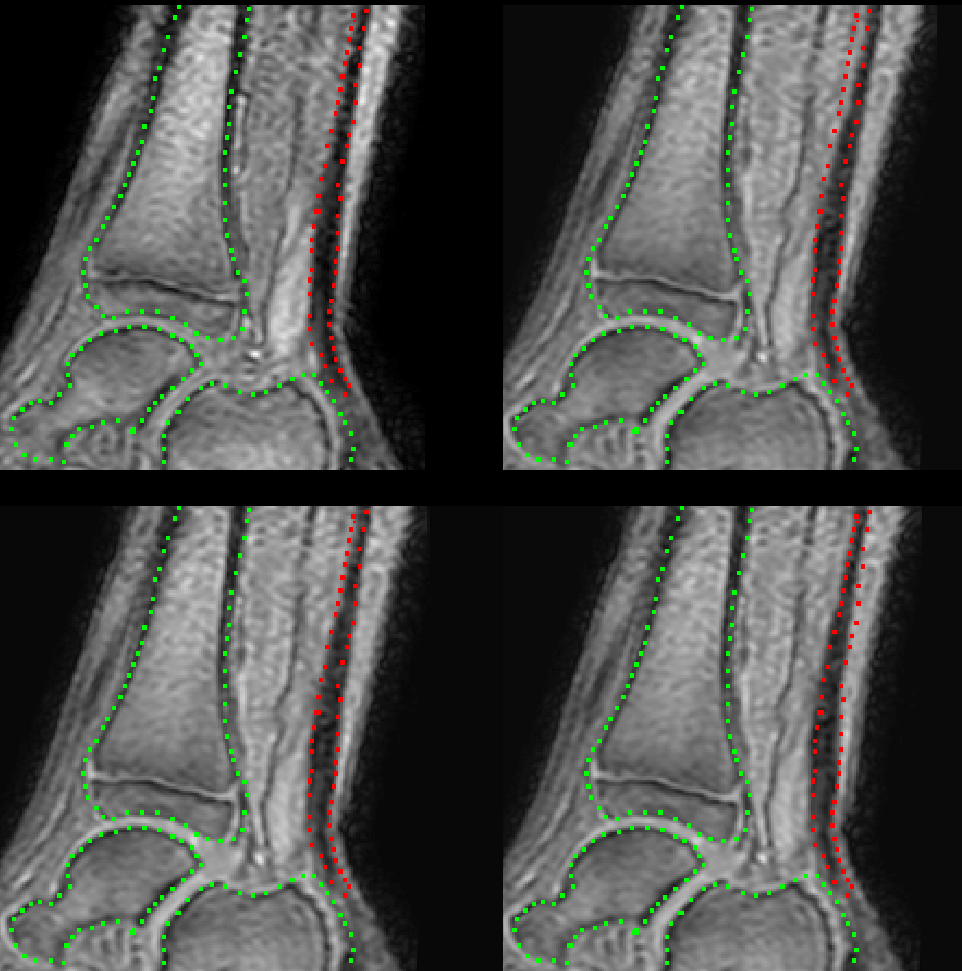}
\caption{\label{fig:deformation_accuracy} Effects of different weight functions to the estimation of deformation fields. From left to right, from up to down: target image; image reconstructed using ~\eqref{eq:weights111} with $\alpha = 0.5$ and $\beta = 1$; image reconstructed using ~\eqref{eq:weights111} with $\alpha = 0.5$ and $\beta = 2$; image reconstructed using eq~\eqref{eq:weights222} with $\gamma = 0.1$. All contour points are drawn in the target image.}

\end{figure}

In this work, a motion-interpolation-based method for increasing the temporal resolution of anatomical dynamic MRI sequences is presented.
We have generalized the LEPF to dynamic articulated structures and we have also proposed new weight functions to increase the local effect of bone transformations on the deformations of the surrounding tissues. The proposed inverse-distance-based weight functions are more suitable than the Gaussian weight functions~\cite{seiler2012capturing} for articulated registration. Since the Gaussian weight functions induces some smoothing which affects the bone shapes by affecting sharp peaks particularly. Both our new weight functions and the weight functions defined in~\cite{commowick2006efficient,commowick2008efficient} conserve the bone topologies. However, our weight functions yield more accurate deformation vectors outside the skeleton as illustrated in Figure~\ref{fig:deformation_accuracy} for the Achilles tendon deformations.

The proposed post processing pipeline aims to overcome the physical limitations related to real-time dynamic MR imaging algorithms which are generally based on compressed sensing theory~\cite{lustig2008compressed}, for which it is hard to fastly acquire the entire or nearly the entire joint trajectory inside the MR scanner because of the limited \textit{k}-space sampling.

\section{Conclusion}
In this paper, we proposed a fast and efficient algorithm to compute the exponential map for solving ODEs modelizing the evolution of dynamical systems over a regular grid. An objective comparison between the eigendecomposition method and the scaling and squaring technique used in previous studies is performed using simulated data and four dimensional (3D + time) dynamic MRI data. Experiments show that the scaling and squaring method is more suitable for sparse grids since computational burden and memory requirements grow exponentially with grid size. The obtained results demonstrate also the potential of the eigendecomposition method to optimally balance the trade-off between accuracies, computation times, and memory requirements for the computation of the exponential map.

\begin{acknowledgements}
This work was supported by the French region of Brittany and the chaire d'excellence INSERM, IMT Atlantique. The authors wish to express their thanks to prof. Nicholas J.Higham for feedback and for valuable comments about this work. 
\end{acknowledgements}
$^*$ The source code is available on github: \url{https://github.com/rousseau/dynMRI/blob/master/Exponential_map.py}

\begin{subappendices}

\subsection{Appendix: numerical example:}
\label{appendix1}
%\subsubsection{Diagonalizability of bone rigid transformations: numerical example:}
In this appendix, we provide a numerical example of the computation of eigenvalues and eigenvectors of realistic bone rigid transforms. Let $T_0$, $T_1$ and $T_2$ be the estimated rigid transformations between two successive acquired time frames, for calcaneus, talus, and tibia, respectively.

\begin{align}
T_0 &= \begin{bmatrix*}
  \mathmakebox[\cellwidth]{0.9941718578} & \mathmakebox[\cellwidth]{0.1057560965} & \mathmakebox[\cellwidth]{0.02092652582} & \mathmakebox[\cellwidth]{-10.089325} \\
  -0.1060616449 & 0.9942600131 & 0.01407066546 & 11.20823699 \\
  -0.0193183478 & -0.01620816253 & 0.9996820092 & 3.384391194 \\
  0 & 0 & 0 & 1
\end{bmatrix*} \\
T_1 &= \begin{bmatrix*}
  \mathmakebox[\cellwidth]{0.9969449639} & \mathmakebox[\cellwidth]{0.07495416701}
    & \mathmakebox[\cellwidth]{0.02196962386} & \mathmakebox[\cellwidth]{-7.610509439} \\
  -0.0742572844 & 0.9967576861 &  -0.03098434582 &  8.603319055 \\
  -0.02422079816 & 0.02925828099 & 0.9992784262 & 0.1743830604 \\
  0 & 0 & 0 & 1
\end{bmatrix*} \\
T_2 &= \begin{bmatrix*}
  \mathmakebox[\cellwidth]{0.9999853969} & \mathmakebox[\cellwidth]{-0.002367701847}
    & \mathmakebox[\cellwidth]{-0.004865686409} & \mathmakebox[\cellwidth]{-0.5515243692} \\
  0.002382844221 & 0.9999923706 & 0.003108616918 & 0.1245495693 \\
  0.004858288914 & -0.003120165784 & 0.9999833703 & 0.09221866638  \\
  0 & 0 & 0 & 1
\end{bmatrix*} 
\end{align}

From Table~\ref{tab:exp_res}, it is clear that $T_0$ and $T_1$ possess four distinct eigenvalues since $\lambda_1 = \lambda_2^*$ and $\lambda_3 \neq \lambda_4=1$. However, $T_2$ has a repeated eigenvalue $\lambda_3 = \lambda_4=1$ (which was expected since $T_2$ is very close to the identity so that all its eigenvalues should be equal or nearly equal to $1$). All bones typically rotate about the three axes to maintain joint stability. For example, the tibial motion is not a pure screw motion and the matrix $T_2$ can be diagonalized as demonstrated in Section~\ref{diagonalizability}, and it has the following eigenvectors:

\[
v_1 = (0.12 + i 0.6;\quad 0.197-i 0.36;\quad 0.67;\quad 0),\]
\[
v_2 = (0.12 - i 0.6;\quad 0.197+i 0.36;\quad 0.67;\quad 0),\]
\[
v_3 = (-0.49;\quad -0.81;\quad 0.32;\quad 0),\] and, 

\[
\tilde{v}_4 = (0.49;\quad 0.81;\quad -0.32;\quad \epsilon = 1.13*10^{-15}).\]

Then, $v_1$, $v_2$, $v_3$, and $\tilde{v}_4$ form an eigenbasis of $\mathbb{C}^4$ and $T_2$ is also diagonalizable despite the fact
that it has a repeated eigenvalue.

\begin{center}
\begin{table}
 \caption{\label{tab:exp_res} experimental results.}
\begin{tabular}{ |c|c|c|c| } 
 \hline
 Matrix & Rotations(deg) & Translations(mm) & $\lambda_3$ \\
 \hline
 $T_0$ & $(-0.802, 1.203, -6.087)$ & $(-10.09, 11.21, 3.38)$ & $1.00030947+0.i$\\ 
 \hline
 $T_1$ & $(1.777, 1.261, -4.302)$ & $(-7.6, 8.603, 0.174)$ & $0.99982089+0.i$\\ 
 \hline
 $T_2$ & $(-0.172, -0.286, 0.114)$ & $(-0.551, 0.1245, 0.092)$ & $1+0.i$\\ 
 \hline
 
\end{tabular}
\end{table}
\end{center}

Table~\ref{tab:exp_res} illustrates the fact that each bones rotates simultaneously about the three axes, which confirms our hypothesis suggesting that the human bones may never be able to perform pure screw transformations under the joint mechanical constraints.

Let now $V = a \quad log(T_0) + b \quad log(T_1) + c \quad log (T_2)$ be a velocity. Then for: $a=0.3$, $b=0.2$, and $c=0.5$, the complex eigenvalues of $V$ are $\{\lambda_1 = -2,9.10^{-5} -i \quad 4,6.10^{-2}; \lambda_2 = -2,9.10^{-5} +i \quad 4,6.10^{-2}$ 
$; \lambda_3 = 4,6.10^{-5} -i \quad 7,5.10^{-28};   \lambda_4 = 0 \}$. These eigenvalues are distincts and consequently $V$ is diagonalizable.

% \begin{bmatrix}
%   \mathmakebox[\cellwidth]{ -3.288e^{-5} -1.635e-22j} & \mathmakebox[\cellwidth]{0} & \mathmakebox[\cellwidth]{0} & \mathmakebox[\cellwidth]{\Delta _{x}}\\
%   0 & 1 & 0 & \Delta _{y}\\
%   0 & 0 & 1 & \Delta _{z}\\
%   0 & 0 & 0 & 1
% \end{bmatrix}
% \end{align}
\end{subappendices}

\bibliography{references.bib}

\bibliographystyle{spmpsci}      % mathematics and physical sciences

\end{document}